\documentclass[conference]{IEEEtran}
\usepackage[utf8]{inputenc}
\usepackage[cmex10]{amsmath}
\usepackage{subfigure}
\usepackage{booktabs}
\usepackage{graphicx}
\usepackage{amsfonts}
\usepackage{flushend} 
\usepackage{dblfloatfix} 

\IEEEoverridecommandlockouts

\date{}
\title{Measuring the Directional Distance Between \\Fuzzy Sets \thanks{This work was partially funded by the EPSRC’s Towards Data-Driven Environmental Policy Design grant, EP/K012479/1 and the RCUK’s Horizon Digital Economy Research Hub grant, EP/G065802/1.}}
\author{
  \IEEEauthorblockN{Josie McCulloch}
  \IEEEauthorblockA{
    School of Computer Science\\
    University of Nottingham\\
    Nottingham, United Kingdom\\
    psxjm5@nottingham.ac.uk}
\and
  \IEEEauthorblockN{Christian Wagner}
  \IEEEauthorblockA{
    School of Computer Science\\
    University of Nottingham\\
    Nottingham, United Kingdom\\
    christian.wagner@nottingham.ac.uk}
\and
  \IEEEauthorblockN{Uwe Aickelin}
  \IEEEauthorblockA{School of Computer Science\\
  University of Nottingham\\
  Nottingham, United Kingdom\\
  uwe.aickelin@nottingham.ac.uk}
}

\begin{document}
\maketitle
\begin{abstract}
The measure of distance between two fuzzy sets is a fundamental tool within fuzzy set theory. However, current distance measures within the literature do not account for the direction of change between fuzzy sets; a useful concept in a variety of applications, such as Computing With Words. In this paper, we highlight this utility and introduce a distance measure which takes the direction between sets into account. We provide details of its application for normal and non-normal, as well as convex and non-convex fuzzy sets. We demonstrate the new distance measure using real data from the MovieLens dataset and establish the benefits of measuring the direction between fuzzy sets.
\end{abstract}

\begin{IEEEkeywords}
 distance measure, fuzzy sets, Hausdorff metric, directional distance
\end{IEEEkeywords}

\section{Introduction} 
\label{sec:introduction}
Distance measures for fuzzy sets are an important tool and have been applied to many fields. For example, Bonissone \cite{BonissoneLinguisticApproach80} illustrates examples of applying distance measures in decision analysis and artificial intelligence and Wang and Xing \cite{Wang20052063} demonstrate distance measures applied to pattern recognition, particularly to the problem of classification. Turksen and Zhang \cite{analogicalReasoning} also demonstrate the applicability of similarity based on distance measures in fuzzy logic inference based on analogical reasoning.

A function $d(A,B) \rightarrow \mathbb{R}^+ $, for which $A$ and $B$ are fuzzy sets in the universe of discourse $X$, is commonly called a distance measure if it satisfies the following properties \cite{Xuecheng1992305}:
\begin{enumerate}
 \item $d(A,B) = d(B,A)$
 \item $d(A,A) = 0$
 \item $d(D, D^c) = max_{A,B \in X}d(A,B)$
 \item If $A \subset B \subset C$, then $d(A,B) \leq d(A,C)$ and $d(B,C) \leq d(A,C)$
\end{enumerate}

Distance measures that are currently in the literature do not account for the ``change in direction'' between fuzzy sets. That is, they reveal the distance between two fuzzy sets, but they do not indicate if a fuzzy set is placed to the left or right of another fuzzy set; a concept which will prove useful within Computing with Words (CW) and the ranking of fuzzy numbers. This paper discusses the value of using a distance measure which identifies the direction of change between fuzzy sets, and proposes an extension of the Hausdorff metric to implement it. Additional extensions of the distance measure are also presented to solve measuring the distance between non-normal and non-convex fuzzy sets. 

In Section \ref{sec:background} we present some background information of fuzzy sets, followed by a discussion presenting the importance of using $\alpha$-cuts in distance measures of fuzzy sets as well as introducing a number of alpha-cut based distance measures currently used in the literature. Section \ref{sec:directional_measures} introduces a new direction-based distance measure, followed by demonstrations of the new measure compared with current measures in Section \ref{sec:directional_experiments}. Demonstrations using synthetic fuzzy sets and real data are presented. Sections \ref{sec:non_normalised} and \ref{sec:non_convex} look at extensions of the newly proposed distance measure for non-normal and non-convex fuzzy sets, respectively. Finally, conclusions are presented in Section \ref{sec:conclusions}.


\section{Background}
\label{sec:background}

\subsection{Fuzzy Sets}
\label{sec:fuzzy_sets}
Fuzzy sets were first introduced by Zadeh \cite{Zadeh1965} in 1965 and have since been applied to many fields, including data mining \cite{datamining}, time-series prediction \cite{timeSeries} and CW \cite{CWW}. A fuzzy set is a set in which the membership of each element is no longer a Boolean, i.e. not 0 or 1, but instead its membership lies in the interval [0,1]. A fuzzy set $F$ may be viewed as a set of ordered pairs as follows \cite{mendel2001uncertain}:
\begin{equation}
 F = (x, \mu_F(x))\ |\ x \in X
\end{equation}
where $\mu_F(x)$ indicates the membership grade of the element $x$ in the fuzzy set $F$. In a discrete universe of discourse, the fuzzy set $F$ can also be written as \cite{mendel2001uncertain}
\begin{equation}
 F = \sum_x \mu_F(x)\ /\ x
\end{equation}
where $\sum$ denotes the collection of all points $x \in X$ with associated membership value $\mu_F(x)$.


\subsection{$\alpha$-cuts vs Vertical Slices}
\label{sec:h_vs_v}
In \cite{Zwick1987221} it is noted that a distance measure for fuzzy sets should ideally focus on the ordering within the $x$-axis as it is this axis that holds the important information regarding where a set's membership lies. 

\begin{figure}
  \centering
  \subfigure[]
  {
      \includegraphics[scale=0.4]{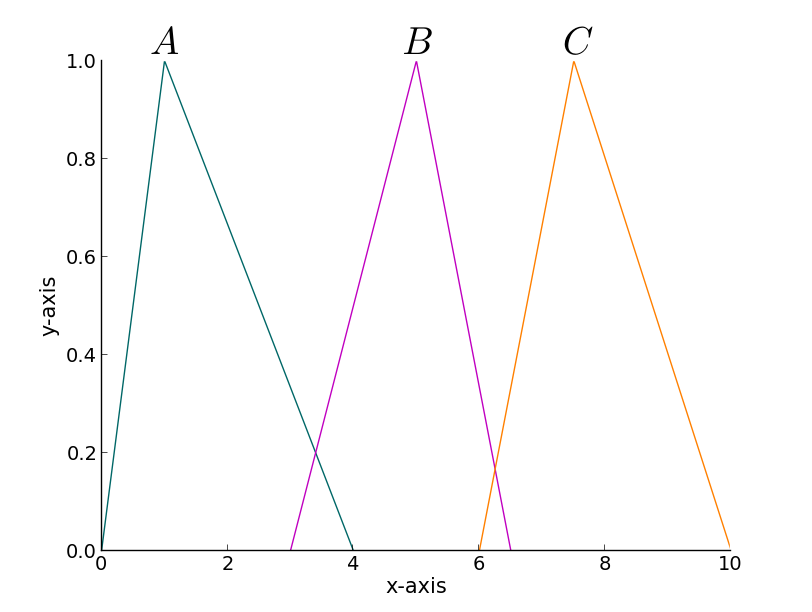}
      \label{fig:different_cuts_plain_sets}
  }
  \subfigure[]
  {
      \includegraphics[scale=0.2]{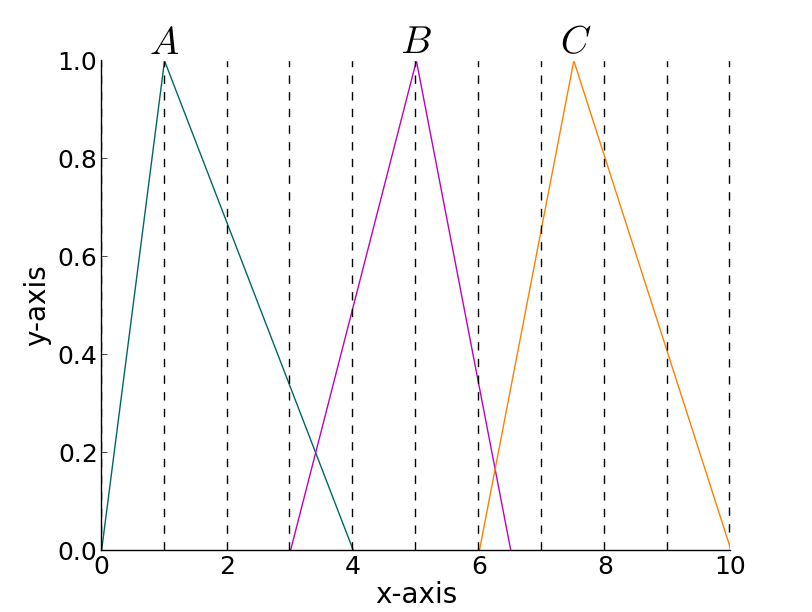}
      \label{fig:example_of_vlines}
  }
  \subfigure[]
  {
    \includegraphics[scale=0.2]{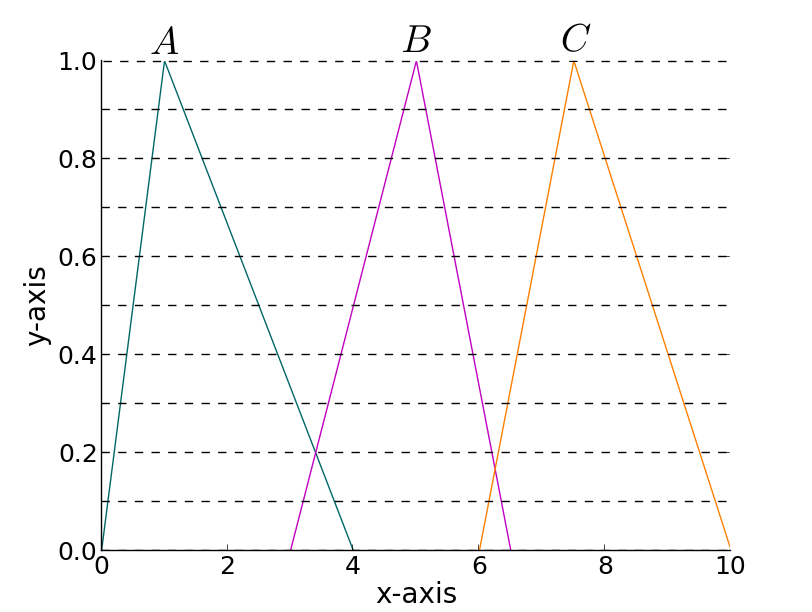}
    \label{fig:example_of_hlines}
  }
  \caption{(a) Three fuzzy sets to demonstrate vertical slices and $\alpha$-cuts; A, B and C
	  (b) Dashed lines representing vertical slices.
	  (c) Dashed lines representing $\alpha$-cuts.}
  \label{fig:different_cuts_example}
\end{figure}

To show the importance of $\alpha$-cuts in distance measures, consider the following example. A survey of three restaurants ($A$, $B$ and $C$) is taken to find out how delicious the food is at each restaurant, on a scale of 0 to 10. Fuzzy sets are then constructed using the results of the survey; Fig. \ref{fig:different_cuts_plain_sets} shows an example of such fuzzy sets. A distance measure can be used on these fuzzy sets to determine how much more delicious the food of one restaurant is compared to another. Two common methods of measuring the distance (or similarity) between fuzzy sets are vertical slices (shown in Fig. \ref{fig:example_of_vlines}) and $\alpha$-cuts (shown in Fig. \ref{fig:example_of_hlines}). The $\alpha$-cut of the fuzzy set $A$ is a non-fuzzy set comprised of all the elements whose membership grade within $A$ is greater than or equal to $\alpha$ \cite{Zadeh1975199}; this is written as $A_\alpha = \{x\ |\ \mu_A(x) \geq \alpha \}$. 

Using a vertical slice approach will give an indication of how much the two sets overlap, capturing the respective food quality. This is useful and most often used when measuring the similarity between the two sets e.g. their intersection, but it does not indicate how far apart they are along the $x$-axis. For example, if the intersection between two fuzzy sets is the empty set then both sets could be infinitely far apart or right next to each other. 

By using $\alpha$-cuts to measure the distance between fuzzy sets we get an approximation of how far apart the two sets are in terms of their universe of discourse. Thus, the result of the distance measure will be a value that is meaningful to the user. For example, if the distance between two of the restaurants is 5 then the user can understand that one of the restaurants was rated approximately 5 points higher than the other restaurant.

Two equations ((\ref{eq:vertical}) and (\ref{eq:horizontal})) have been created to demonstrate this idea. (\ref{eq:vertical}) uses vertical slices and is written as follows:
\begin{equation}
 d(A,B) = \frac{1}{n} \sum^n_{i=1} | \mu_A(x_i) - \mu_B(x_i) |
 \label{eq:vertical}
\end{equation}
where $n$ is the total number of discretisations on the $x$-axis. The next equation (\ref{eq:horizontal}) uses $\alpha$-cuts and is written as:
\begin{equation}
 d(A,B) = \frac{1}{m} \sum^m_{i=0} 
      max \{ | A_{\alpha_{i_l}} - B_{\alpha_{i_l}} |, | A_{\alpha_{i_r}} - B_{\alpha_{i_r}} | \}
 \label{eq:horizontal}
\end{equation}
where $A_{\alpha_{i_l}}$ is the left point of the $\alpha$-cut of $A$ at $\alpha_i$, $A_{\alpha_{i_r}}$ is the right point of the $\alpha$-cut of $A$ at $\alpha_i$, and $m$ is the total number of $\alpha$-cuts. The fuzzy sets used in this demonstration are shown in Fig. \ref{fig:different_cuts_plain_sets}. Both the $x$-axis and $y$-axis were discretised into points each at a distance of $0.1$. The results of this demonstration are displayed in Table \ref{tab:slices_demonstration}.
Examining the fuzzy sets in Fig. \ref{fig:different_cuts_plain_sets}, it would be sensible to describe set $C$ as being at a greater distance from set $A$ than set $B$ is from $A$. However, it can be clearly seen that when using vertical slices the result of the distance measure decreases as the sets are placed further apart along the $x$-axis. This is a result of the equation measuring the distance between the fuzzy sets according to their membership values on the $y$-axis, which can lead to unintuitive results when the fuzzy sets being measured are disjoint. For example, between $x=4$ and $x=6$ the distance between sets $A$ and $C$ is $0$ because they both have the same membership value at these coordinates. 

Considering the $\alpha$-cut approach, the distance according to the measure increases as the sets move further apart along the $x$-axis. This seems logical as it is the $x$-axis that is of most importance when defining sets so it is generally this axis that a user is interested in when comparing different fuzzy sets.

\setlength{\tabcolsep}{6pt}
\begin{table}
\caption{A comparison of vertical slice and $\alpha$-cut approaches on measuring distance using the fuzzy sets in Fig. \ref{fig:different_cuts_plain_sets}.}
  \begin{center}
    \begin{tabular}{  c  c  c  c  }
      \toprule
        & $(A, A)$ & $(A, B)$ & $(A, C)$ \\ \midrule
        Vertical slice based distance (\ref{eq:vertical}) & 0.0 & 0.538 & 0.396 \\
        $\alpha$-cut based distance (\ref{eq:horizontal}) & 0.0 & 3.495 & 6.262
        \\ \bottomrule
    \end{tabular}
  \end{center}
  \label{tab:slices_demonstration}
\end{table}


\subsection{Current Distance Measures}
\label{sec:current_measures}
In this section, the Hausdorff metric for two intervals is reviewed, followed by two existing distance measures for fuzzy sets which use $\alpha$-cuts. 

The Hausdorff metric gives a generalisation of the distance between two non-empty crisp sets. The Hausdorff distance between two intervals is defined as \cite{Zwick1987221}:
\begin{equation}
 h(\bar{A}, \bar{B}) = max \{| \bar{A}_l - \bar{B}_l |, | \bar{A}_r - \bar{B}_r |\}
 \label{eq:interval_haus}
\end{equation}
where $\bar{A} = [\bar{A}_l, \bar{A}_r]$ and $\bar{B} = [\bar{B}_l, \bar{B}_r]$. This is a common metric used to measure the distance between two $\alpha$-cuts. Two common distance measures for fuzzy sets which use the Hausdorff metric are introduced next. 

Ralescu and Ralescu \cite{Ralescu198485} introduced the following generalisation of the Hausdorff metric to measure the distance between fuzzy sets:

\begin{equation}
 d_{RR}(A,B) = \int^1_{\alpha=0} h(A_\alpha, B_\alpha) d \alpha
 \label{eq:ralescu}
\end{equation}
where $h$ is the conventional Hausdorff metric as shown in (\ref{eq:interval_haus}).

Chaudhuri and Rosenfeld also proposed a new metric to determine the distance between fuzzy sets based on the Hausdorff metric as follows \cite{Chaudhur19961157}.

\begin{equation}
d_{CR}(A, B) = \frac{\sum^m_{\alpha=1} y_\alpha\ h(A_\alpha, B_\alpha)}{\sum^m_{\alpha=1} y_\alpha}
\label{eq:CR_haus}
\end{equation}
where the $y$-axis is discretised into $m$ points ($y_1, y_2, ..., y_m$), $A_\alpha$ is the non-fuzzy $\alpha$-cut set of the fuzzy set $A$ at y-coordinate $y_\alpha$, and $h$ is the conventional Hausdorff metric in (\ref{eq:interval_haus}). 

Both Ralescu \& Ralescu's and Chaudhuri \& Rosenfeld's distance measures assume the fuzzy sets being measured are normal (i.e. $\exists x \in X\ \mu_A(x)=1$ and $\exists x \in X\ \mu_B(x)=1$). However, an extension of Chaudhuri \& Rosenfeld's measure (\ref{eq:CR_haus}) for non-normal fuzzy sets (i.e. $\forall x \in X\ \mu_A(x) < 1$ or $\forall x \in X\ \mu_B(x) < 1$) is given in \cite{Chaudhur19961157} and is discussed in Section \ref{sec:non_normalised}.

Having reviewed the core concepts of this paper, we proceed to develop direction-based distance measures for normal and convex, as well as non-normal and non-convex fuzzy sets.


\begin{table*}[b!]
  \caption{Results of distance measure applied to the fuzzy sets in Fig. \ref{fig:type1_exp}.}
  \begin{center}
    \begin{tabular}{  c  c  c  c  c  c  c  c  c  c  }
      \toprule
        & $(A, A)$ & $(A, B)$ & $(A, C)$ & $(C, A)$ & $(B, A)$ & 
          $(A, D)$ & $(D, A)$ & $(A, E)$ & $(E, A)$\\ \midrule
          Ralescu \& Ralescu  \& (\ref{eq:interval_haus}) & 
            0.0 & 3.0 & 6.0 & 6.0 & 3.0 & 4.0 & 4.0 & 2.0 & 2.0 \\
	  Ralescu \& Ralescu  \& (\ref{eq:interval_haus_with_sign}) & 
            0.0 & 3.0 & 6.0 & -6.0 & -3.0 & 4.0 & -4.0 & 2.0 & -2.0 \\
          Chaudhuri \& Rosenfeld \& (\ref{eq:interval_haus}) &
            0.0 & 3.0 & 6.0 & 6.0 & 3.0 & 3.65 & 3.65 & 1.65 & 1.65 \\
          Chaudhuri \& Rosenfeld \& (\ref{eq:interval_haus_with_sign}) &
            0.0 & 3.0 & 6.0 & -6.0 & -3.0 & 3.65 & -3.65 & 1.65 & -1.65
        \\ \bottomrule
    \end{tabular}
  \end{center}
  \label{tab:results}
\end{table*}

\section{Directional Distance Measures}
\label{sec:directional_measures}
The following proposes the idea that the property of symmetry for distance measures (i.e., $d(A,B) = d(B,A)$) is not ideal for every problem. For example, consider two fuzzy sets that represent on a scale of 1 to 10 how fun two different roller coasters are according to a public survey (1 denoting not at all and 10 meaning very much). When comparing the distance between these fuzzy sets, the result will indicate how similar the roller coasters are in terms of how fun they are, but it will not indicate which is more fun. For example, if $d(\text{rollercoasterA}, \text{rollercoasterB}) = 6$, we will know that one of them was given, roughly speaking, ``6 more points'' than the other. However, there is no way to determine which is the more fun roller coaster without visually checking which set is actually on the left or right of the other. This can be time-consuming and tedious if many comparisons need to be made. 

Instead, it would be ideal if the result were a signed value. For example, the result will be a positive value if rollercoasterB is more fun than rollercoasterA, and will be a negative value if the opposite case is true. For example $d(\text{rollercoasterA}, \text{rollercoasterB}) = 6$ indicates that rollercoasterB was rated approximately 6 points higher than rollercoasterA. For the same case, it will be true that $d(\text{rollercoasterB}, \text{rollercoasterA}) = -6$, indicating that rollercoasterA was rated approximately 6 points less than rollercoasterB. Now, it is clear from the result of $d$ which roller coaster is the most fun and by how much.

By taking this approach, the distance measure will no longer have the property of symmetry (i.e., $d(A,B) \neq d(B,A)$), however the absolute values of $d(A,B)$ and $d(B,A)$ will be equal (i.e. $|d(A,B)| = |d(B,A)|$). 

The concept of a directional distance measure will prove useful in analysing survey data as shown in Section \ref{sec:movielens} and will be a valuable tool in CW \cite{CWW} and, more generally, in the evaluation of ratings and rankings of fuzzy numbers and sets.

As mentioned in (\ref{eq:interval_haus}), when comparing intervals, the Hausdorff metric is described by
\begin{equation}
  \nonumber
  h(\bar{A}, \bar{B}) = max \{|\bar{A}_l - \bar{B}_l |, |\bar{A}_r - \bar{B}_r |\}
\end{equation}
where $ \bar{A} = [\bar{A}_l, \bar{A}_r] $ and $ \bar{B} = [\bar{B}_l, \bar{B}_r] $ \cite{Zwick1987221}. 
Currently, this will never give a negative value for a negative distance. However, it can be modified as follows:

\begin{equation}
  h(\bar{A}, \bar{B})=
  \begin{cases}
    \bar{B}_l - \bar{A}_l, & \text{if $|\bar{B}_l - \bar{A}_l| > |\bar{B}_r - \bar{A}_r| $}.\\
    \bar{B}_r - \bar{A}_r, & \text{otherwise}.
  \end{cases}
  \label{eq:interval_haus_with_sign}
\end{equation}

\begin{figure}[h!]
  \centering
  \includegraphics[scale=0.5]{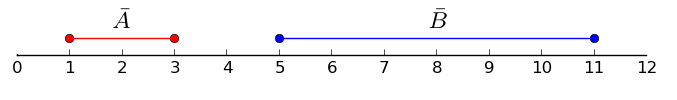}
  \caption{Two interval sets, $\bar{A}$ and $\bar{B}$.}
  \label{fig:intervals}
\end{figure}

This ensures that both the maximum distance and the sign are preserved. For example, in Fig. \ref{fig:intervals}, set $\bar{A}$ is defined as $[1,3]$ and set $\bar{B}$ is defined as $[5,11]$. By using (\ref{eq:interval_haus_with_sign}) for $h(\bar{A}, \bar{B})$ the result of $h$ is $8$. Alternatively, when computing $h(\bar{B}, \bar{A})$ the result of $h$ is $-8$. 

By testing $|\bar{B}_l - \bar{A}_l| > |\bar{B}_r - \bar{A}_r|$ within (\ref{eq:interval_haus_with_sign}), the absolute values of $h(\bar{A}, \bar{B})$ and $h(\bar{B}, \bar{A})$ remain the same as when using the conventional Hausdorff metric (\ref{eq:interval_haus}). It also follows that the modified property $|d(A,B)| = |d(B,A)|$ is satisfied.

\section{Demonstrations}
\label{sec:directional_experiments}
This section presents demonstrations of the new direction-based distance measure. The distance measures by Ralescu and Ralescu given in (\ref{eq:ralescu}) and Chaudhuri and Rosenfeld (\ref{eq:CR_haus}) are presented using the regular Hausdorff metric for intervals as shown in (\ref{eq:interval_haus}), and the new direction-based measure shown in (\ref{eq:interval_haus_with_sign}).
Two demonstrations are presented in this section. The first uses synthetic fuzzy sets, and the second uses fuzzy sets constructed from real data.

\subsection{Synthetic Example}
\label{sec:directional_experiments_synthetic}
\begin{figure}
  \centering
  \subfigure[]
  {
      \includegraphics[scale=0.3]{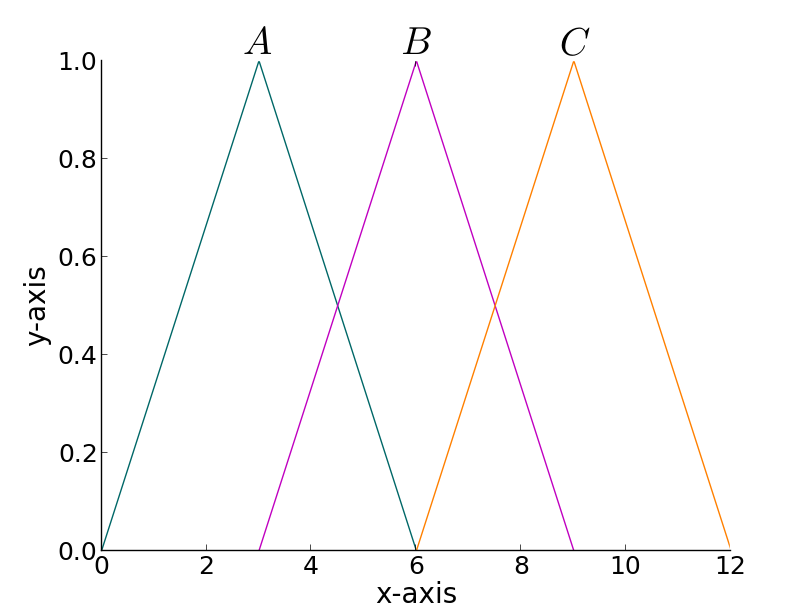}
      \label{fig:type1_sets_ABC}
  }
  \subfigure[]
  {
    \includegraphics[scale=0.3]{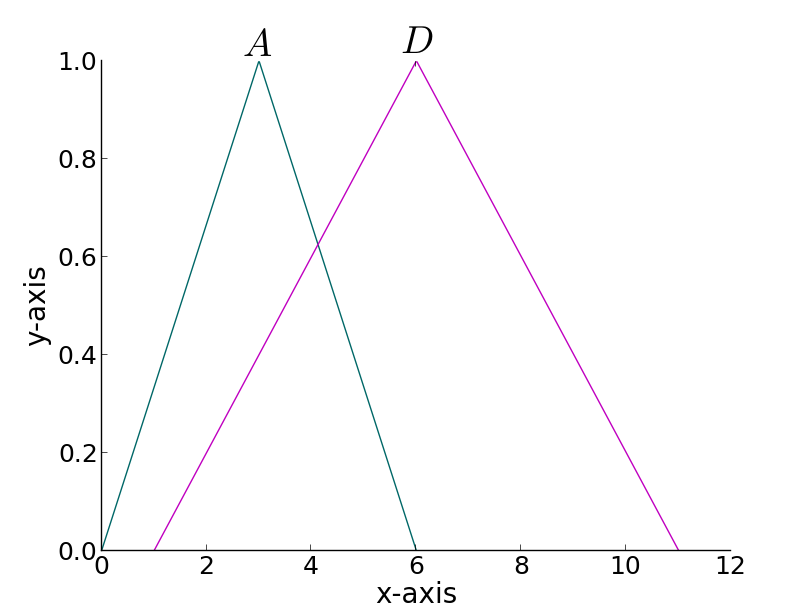}
    \label{fig:type1_a_and_f}
  }
  \subfigure[]
  {
    \includegraphics[scale=0.3]{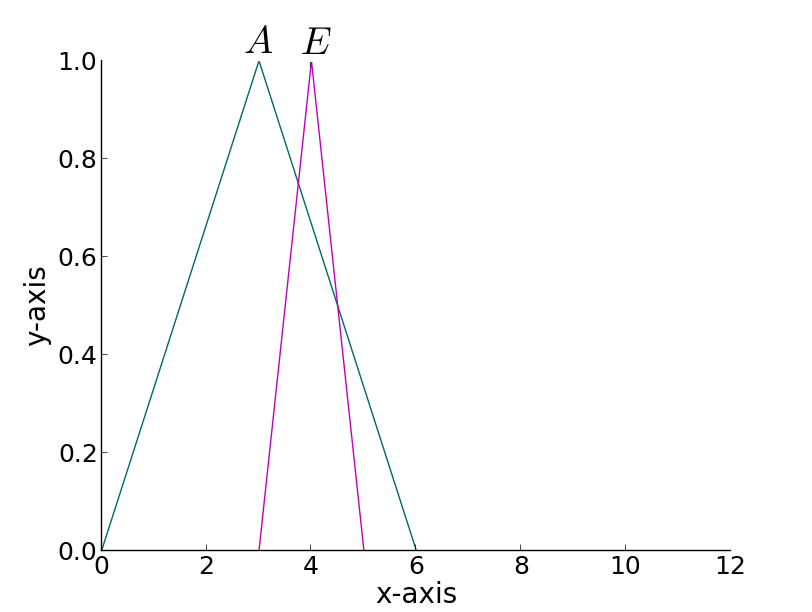}
    \label{fig:type1_a_and_h}
  }
  \caption{Fuzzy sets used to demonstrate the distance measures. 
	   (a) Fuzzy sets A, B and C. 
	   (b) Fuzzy sets A and D. 
	   (c) Fuzzy sets A and E.}
  \label{fig:type1_exp}
\end{figure}

The fuzzy sets shown in Fig. \ref{fig:type1_exp} were constructed to demonstrate the new direction-based distance measure compared to conventional measures. For this demonstration, 51 $\alpha$-cuts were taken for the distance measures. We next present and discuss the results of this demonstration shown in Table \ref{tab:results}. 

It can be seen through the measurements based on sets $A$ through $C$, that as the distance between the sets along the $x$-axis increases, the value of the distance measures also increases. Also, for the proposed direction-based measures based on (\ref{eq:interval_haus_with_sign}), if the fuzzy set given as the second parameter to the distance function is placed to the right of the fuzzy set given as the first parameter then the distance according to the measure is a positive value, and if the fuzzy set of the second parameter is to the left of the fuzzy set of the first parameter then the distance measure gives a negative value. 
Also, it can be seen by the results between fuzzy sets $A$ and $D$ and fuzzy sets $A$ and $E$ that if the peak of the second fuzzy set is positioned to the right of the peak of the first fuzzy set then the result of the distance measure will be a positive value.

Note from Table \ref{tab:results} that when using (\ref{eq:interval_haus_with_sign}), the modified property of distance measures has been obeyed. Only the sign of the value has changed, but their absolute values are the same. For example, $d(A, C)$ and $d(C, A)$ share the same absolute value, it is only their signs that differ. Additionally, it can be seen in Table \ref{tab:results} that the results from Ralescu \& Ralescu's and Chaudhuri \& Rosenfeld's equations share the same absolute value using (\ref{eq:interval_haus_with_sign}) and (\ref{eq:interval_haus}).

\subsection{Real World Example}
\label{sec:movielens}
MovieLens is a dataset of movie ratings developed by the GroupLens project at the University of Minnesota. Their datasets are available at \textit{http://www.grouplens.org}. The dataset used for this demonstration is the 100k MovieLens dataset, which consists of 100,000 ratings from 943 users on 1682 movies, where each rating is between 1 and 5. 

\begin{figure}[t]
  \centering
  \subfigure[]
  {
      \includegraphics[scale=0.25]{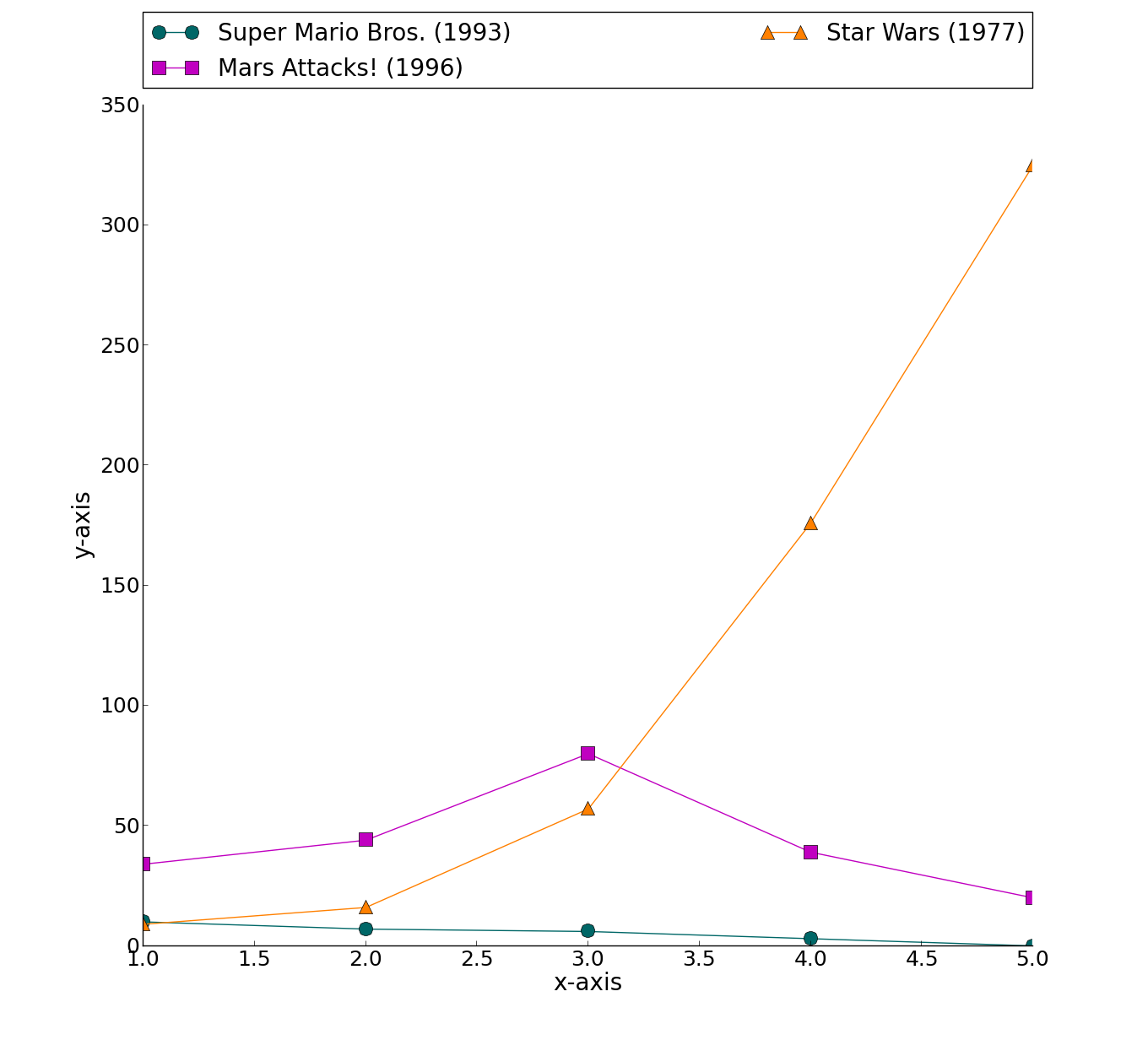}
      \label{fig:films_original}
  }
  \subfigure[]
  {
      \includegraphics[scale=0.26]{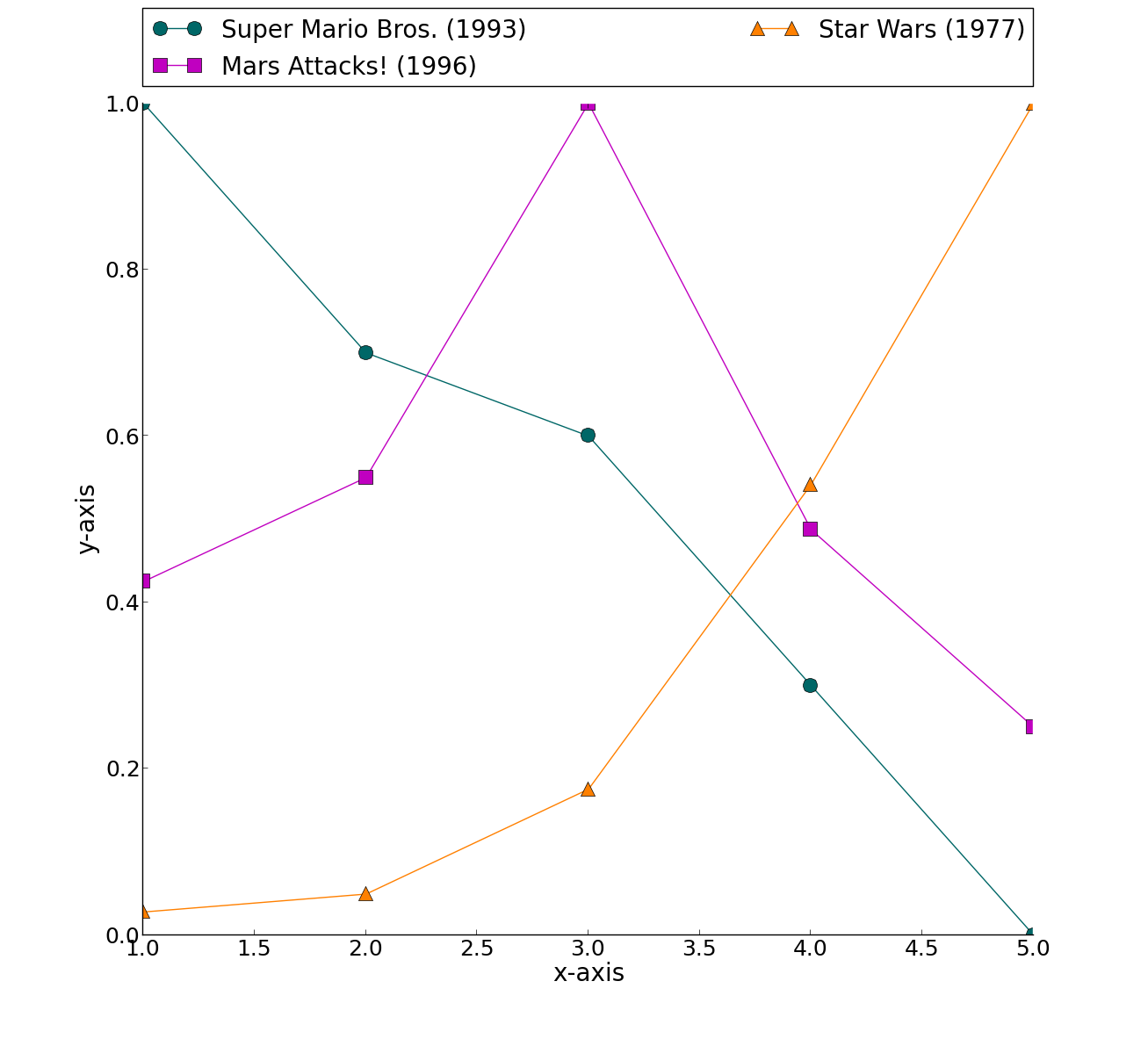}
      \label{fig:films_normalised}
  }
  \caption{(a) Histogram based fuzzy sets of three film ratings from the MovieLens dataset. 
	   (b) Normalised histogram based fuzzy sets of the same films.}
  \label{fig:film_ratings}
\end{figure}

Fuzzy sets were constructed for each film by calculating the histogram of all ratings given to each film, and linear interpolation was used to find the values between known points. The distance measures introduced thus far have been designed only for normalised fuzzy sets, so it is necessary that the histograms of the films are normalised so that the distance measures can be used. The sets were normalised by dividing the $y$ value at each $x$-coordinate by the peak $y$ value; this ensures that the peak value of each fuzzy set is now $1$. The histograms for the three films \textit{Super Mario Bros.} (SMB), \textit{Mars Attacks!} (MA) and \textit{Star Wars} (SW) and their normalised fuzzy sets are shown in Fig. \ref{fig:film_ratings}. The two previously shown distance measures by Ralescu and Ralescu (RR) (\ref{eq:ralescu}) and by Chaudhuri and Rosenfeld (CR) (\ref{eq:CR_haus}) are demonstrated using the conventional Hausdorff metric in (\ref{eq:interval_haus}) and the proposed, modified, direction-based Hausdorff measure in (\ref{eq:interval_haus_with_sign}). These measures were applied to the fuzzy sets in Fig. \ref{fig:films_normalised} to determine how much better or worse each film was rated compared to each other film. The results of this experiment, for which 51 $\alpha$-cuts were taken, are shown in Table \ref{tab:film_results}.

The results of the distance measures shown in Table \ref{tab:film_results} indicate that the difference in rating between MA and SW is slightly greater than the distance between SMB and MA Referring to the fuzzy sets in Fig. \ref{fig:films_normalised}, these results are expected; the membership functions of SMB and MA are much closer to each other compared to the membership functions of MA and SW. Additionally, referring to Fig. \ref{fig:films_normalised}, it is expected that the greatest distance between any two films is between SMB and SW as their membership functions share an inverse relationship, and the peaks of their membership functions are the furthest apart than any other two films. In Table \ref{tab:film_results} it can be seen that SMB and SW do have the greatest distance by a considerable margin.

Observing the measures by Ralescu \& Ralescu and Chaudhuri \& Rosenfeld when using the conventional Hausdorff metric (\ref{eq:interval_haus}), the difference in ratings between each film is clear. For example, in both measures the distance between SMB and SW is clearly greater than the distance between SMB and MA, however, it is not clear which film is better. When using the direction-based Hausdorff measure in (\ref{eq:interval_haus_with_sign}) the latter is not the case. Thus, the extended measure has shown additional information whilst also maintaining the same absolute values that are produced when using the conventional Hausdorff metric.

\setlength{\tabcolsep}{2pt}
\begin{table}
  \scriptsize
  \caption{Results of distance measure applied to the fuzzy sets in Fig. \ref{fig:films_normalised}.}
  \begin{center}
    \begin{tabular}{  c  c  c  c  c  c  c }
      \toprule
        & (SMB, MA) & (SMB, SW) & (MA, SW) & (MA, SMB) & (SW, SMB) &  (SW, MA) \\ \midrule
        RR \& (\ref{eq:interval_haus})     & 1.194  & 2.775  & 1.974   & 1.194   & 2.775   & 1.974    \\
        RR \& (\ref{eq:interval_haus_with_sign}) & 1.194  & 2.775  & 1.974  & -1.194   & -2.775  & -1.974    \\
        CR \& (\ref{eq:interval_haus}) & 1.399 & 3.270 & 2.097 & 1.399  & 3.270  & 2.097  \\
        CR \& (\ref{eq:interval_haus_with_sign}) & 1.399 & 3.270 & 2.097 & -1.399 & -3.270 & -2.097
        \\ \bottomrule
    \end{tabular}
  \end{center}
  \label{tab:film_results}
\end{table}
\normalsize

\section{Non-normal Fuzzy Sets}
\label{sec:non_normalised}
\begin{figure}[t]
  \centering
  \includegraphics[scale=0.25]{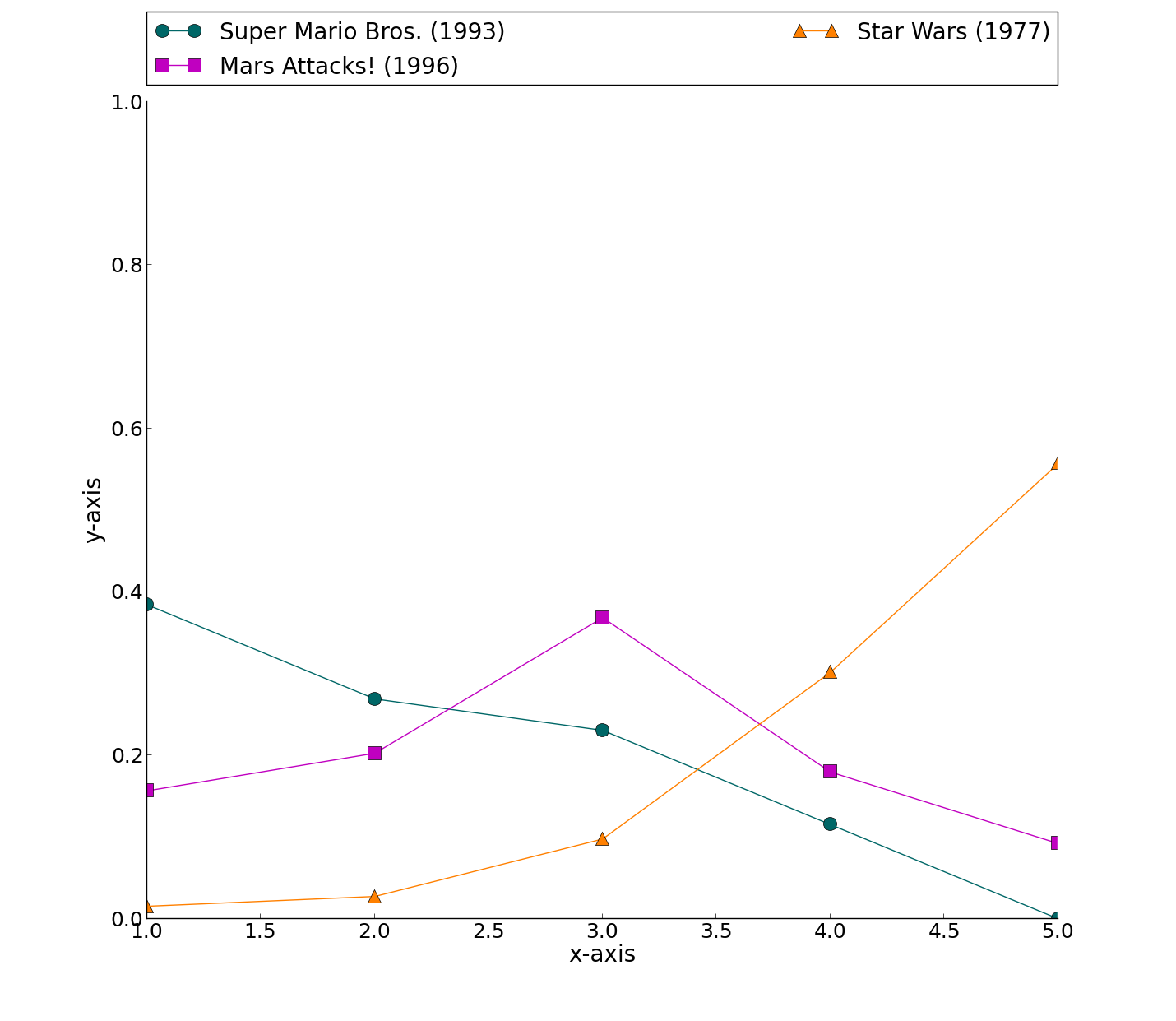}
  \caption{Three non-normal film ratings from the MovieLens dataset.}
  \label{fig:film_ratings_non_normalised}
\end{figure}

Thus far, the distance measures introduced can only be applied to normalised fuzzy sets, where the peak of a set's membership is $1$. However, it is common for sets to be non-normal. Referring to Fig. \ref{fig:film_ratings}, it is clear from the original histograms in Fig. \ref{fig:films_original} that SW received a higher proportion/number of ratings of 5 than MA did of 3, however, this information is lost in the normalised histograms in Fig. \ref{fig:films_normalised}. The following process was performed to retain this information. For every film, each value on the $y$-axis pertaining to the number of times a rating on the $x$-axis was given is divided by the total number of ratings given for that film; ergo the membership value $\mu_A(x)$ now indicates the percentage of people that gave the rating $x$ for the film $A$. This was applied to all of the films in Fig. \ref{fig:films_original} resulting in the fuzzy sets shown in Fig. \ref{fig:film_ratings_non_normalised}.

It can be seen that the current $\alpha$-cut approach to measuring distance cannot be used on these sets. For example, comparing SMB with SW at $\alpha = 0.5$, the $\alpha$-cut of SMB is $[null, null]$ because the set is not present at this $\alpha$-level. However, SW has an $\alpha$-cut of $[4.8, 5.0]$. This poses the question `how can the distance between these two fuzzy sets be measured?' Existing solutions from the literature include the following approaches by Chaudhuri and Rosenfeld \cite{Chaudhur19961157} and Fan \cite{Fan1998793}.

Chaudhuri and Rosenfeld put forward an extension of their distance measure (\ref{eq:CR_haus}) in \cite{Chaudhur19961157}. Firstly, each set is modified so that the maximum membership value of each fuzzy set is $1$, and these modified fuzzy sets are applied to (\ref{eq:CR_haus}). Next, the original, non-modified fuzzy sets are applied to the following equation \cite{Chaudhur19961157}
\begin{equation}
 e(A, B) = \varepsilon \frac{\sum_{x \in X} | \mu_A(x) - \mu_B(x) |}{\sum_{x \in X} x}
 \label{eq:CR_nn_e}
\end{equation}
where $\varepsilon$ is a small positive constant, and its value is determined by the importance of the equation. Finally (\ref{eq:CR_haus}) and (\ref{eq:CR_nn_e}) are joined together as follows \cite{Chaudhur19961157}:
\begin{equation}
  \begin{array}{l}
 d_{CR}(A, B) = \\
    \frac{\sum^m_{\alpha=1} y_\alpha\ h(A_\alpha, B_\alpha)}{\sum^m_{\alpha=1} y_\alpha} 
	      + \varepsilon \frac{\sum_{x \in X} | \mu_A(x) - \mu_B(x) |}{\sum_{x \in X} x}
  \end{array}
  \label{eq:CR_non_normalised}
\end{equation}

Using Chaudhuri and Rosenfeld's method in (\ref{eq:CR_non_normalised}) with the direction-based Hausdorff measure in (\ref{eq:interval_haus_with_sign}), we have altered the extension for non-normal fuzzy sets to no longer take the absolute value $| \mu_A(x_i) - \mu_B(x_i) |$. This is in keeping with the desire to maintain information regarding the direction of change between the fuzzy sets, and ensures that the property $|d(A, B)| = |d(B, A)|$ is maintained. Thus, Chaudhuri and Rosenfeld's measure is altered to 

\begin{equation}
  \begin{array}{l}
 d_{CR}(A, B) = \\
    \frac{\sum^m_{\alpha=1} y_\alpha\ h(A_\alpha, B_\alpha)}{\sum^m_{\alpha=1} y_\alpha} 
	      + \varepsilon \frac{\sum_{x \in X} \mu_B(x) - \mu_A(x) }{\sum_{x \in X} x}
  \end{array}
  \label{eq:CR_non_normalised_altered}
\end{equation}

Fan \cite{Fan1998793} also put forward the following extension of the Hausdorff metric for non-normal fuzzy sets within the compact metric space $S$. Let $h(\emptyset, \emptyset) = 0$ and $h(\emptyset, U) = h(U, \emptyset) = w$ for all non-empty sets $U$, where $w = \sup h(U,V)$ for all non-empty compact subsets $U,V \subset S$. Based on Fan's measure, we propose the following extension. 

Let $h(A_\alpha, \emptyset) = h(A_{\alpha_k}, B_{\alpha_k})$ where $\alpha_k $ is the $\alpha$-level at $max\{ |h(A_\alpha, B_\alpha)| \}\ \forall \alpha\ A_\alpha \neq \emptyset \wedge B_\alpha \neq \emptyset$.
Likewise, let $h(\emptyset, A_\alpha) = h(B_{\alpha_k}, A_{\alpha_k})$ where $\alpha_k $ is the $\alpha$-level at $max\{ |h(B_\alpha, A_\alpha)| \}\ \forall \alpha\ A_\alpha \neq \emptyset \wedge B_\alpha \neq \emptyset$.
By using this, the distance between an $\alpha$-cut and the empty set is the maximum distance between the non-empty $\alpha$-cuts of the sets being measured. Additionally, this approach also ensures that the sign of the distance is maintained.

We propose to disregard $h(\emptyset, \emptyset)$ for the following reason. 
Using (\ref{eq:interval_haus_with_sign}), the distance between SMB and SW in Fig. \ref{fig:film_ratings_non_normalised} at $\alpha = 0.36$ is 2.52. Next, the distance at $\alpha = 0.46$ (at which the $\alpha$-cut of SMB is the empty set) is 2.56. This was calculated using the newly introduced extension based on Fan's extension \cite{Fan1998793}.  Considering this, it is unrealistic to state that the distance at $\alpha = 0.56$, at which the $\alpha$-cut of each film is the empty set, is 0. Though neither fuzzy set is present at this $\alpha$-cut, it is nonsensical to describe the fuzzy sets as having a distance of 0 when all other $\alpha$-cuts denote otherwise. Taking this into account, we propose the following new distance measure for non-normal fuzzy sets:

\begin{equation}
d_{CRF}(A, B) = \frac{
    \sum_{\alpha \in [0, \lambda]]} y_\alpha\ h(A_\alpha, B_\alpha)}
    {\sum_{\alpha \in [0, \lambda]}  y_\alpha }
\label{eq:proposed_non_normalised}
\end{equation}
where $\lambda = \sup\{\alpha \in [0,1]: A_\alpha \neq \emptyset \vee B_\alpha \neq \emptyset\}$ and $h$ is described in (\ref{eq:interval_haus_with_sign}). A numerical example of measuring the distance between non-normal fuzzy sets using (\ref{eq:proposed_non_normalised}) is presented in the appendix.

Demonstrations were performed on these non-normal fuzzy sets using Chaudhuri and Rosenfeld's altered method given in (\ref{eq:CR_non_normalised_altered}), and the proposed method shown in (\ref{eq:proposed_non_normalised}). Both equations use the modified direction-based Hausdorff measure in (\ref{eq:interval_haus_with_sign}). Each axis was discretised into 51 equidistant points, and $\varepsilon$ was set as 1. The results are shown in Table \ref{tab:film_results_non_normalised}.

Comparing the results on the normalised fuzzy sets in Table \ref{tab:film_results} with the results on the non-normal fuzzy sets in Table \ref{tab:film_results_non_normalised}, the distances of the non-normal sets have changed significantly. For example, in Table \ref{tab:film_results_non_normalised}, according to (\ref{eq:proposed_non_normalised}), SMB and SW are now approximately as close to each other as MA and SW. Whereas in the previous experiment in Table \ref{tab:film_results} the films SMB and SW have the greatest distance by a considerable margin. In Fig. \ref{fig:film_ratings_non_normalised}, the difference between the membership functions of SMB and MA is much smaller than in Fig. \ref{fig:film_ratings} because their membership functions have been compressed. It is due to their membership functions being more similar that the distance of these films from SW is also similar. The distances according to (\ref{eq:CR_non_normalised_altered}), however, are approximately the same in both experiments. Note that in Table \ref{tab:film_results_non_normalised} both Chaudhuri \& Rosenfeld's altered measure and the proposed measure based on Fan's extension still satisfy $|d(A,B)| = |d(B,A)|$.

\setlength{\tabcolsep}{1pt}
\begin{table}[t]
  \scriptsize
  \caption{Results of distance measures applied to the fuzzy sets in Fig. \ref{fig:film_ratings_non_normalised}.}
  \begin{center}
    \begin{tabular}{  c  c  c  c  c  c  c }
      \toprule
        & (SMB, MA) & (SMB, SW) & (MA, SW) & (MA, SMB) & (SW, SMB) &  (SW, MA) \\ \midrule
	(\ref{eq:CR_non_normalised_altered}) \& (\ref{eq:interval_haus_with_sign})  
	      & 1.431 & 3.261 & 2.057 & -1.431 & -3.261 & -2.057 \\
        (\ref{eq:proposed_non_normalised}) \& (\ref{eq:interval_haus_with_sign}) 
	      & 0.904 & 2.374 & 2.348 & -0.904 & -2.374 & -2.348
        \\ \bottomrule
    \end{tabular}
  \end{center}
  \label{tab:film_results_non_normalised}
\end{table}
\normalsize

\section{Non-convex Fuzzy Sets}
\label{sec:non_convex}
\begin{figure}[t]
  \centering
  \includegraphics[scale=0.28]{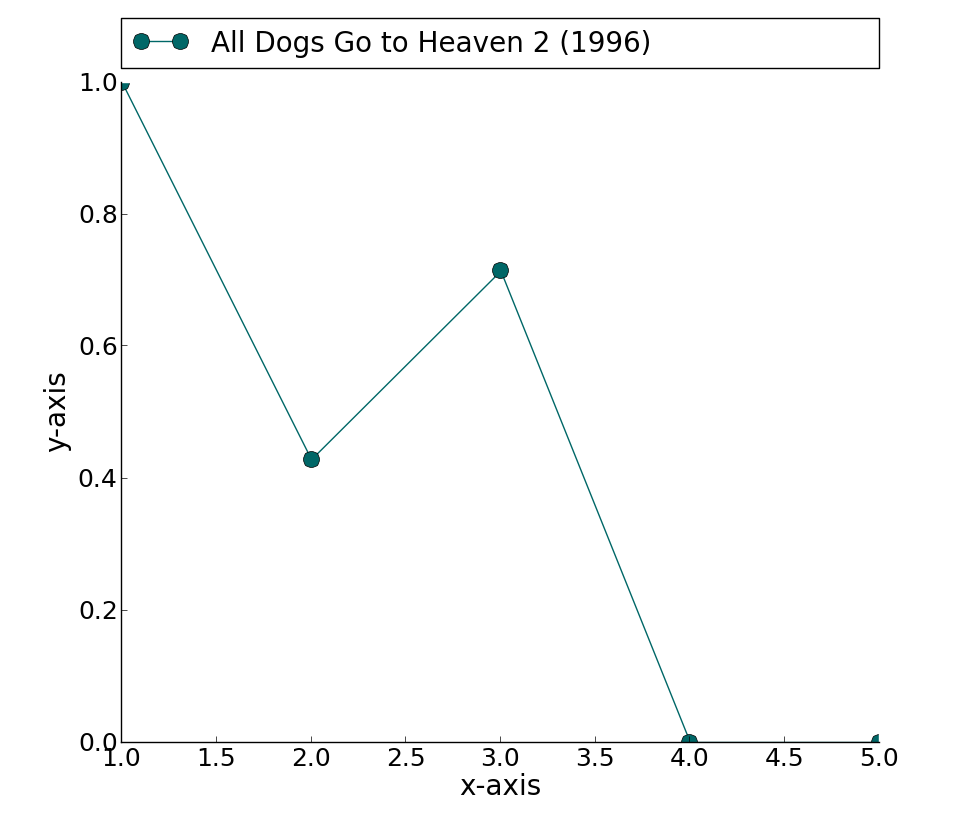}
  \caption{The non-convex fuzzy set for the film \textit{All Dogs Go to Heaven 2}.}
  \label{fig:non_convex_film_set}
\end{figure}

The distance measures discussed so far can only be utilised on convex fuzzy sets. For example, consider the fuzzy set for the film \textit{All Dogs Go to Heaven 2} shown in Fig. \ref{fig:non_convex_film_set}. The $\alpha$-cut for this film at $y=0.6$ yields four points. The Hausdorff measure in (\ref{eq:interval_haus}) and  (\ref{eq:interval_haus_with_sign}) can so far only be applied when there are two distinct end points, and therefore cannot be applied to this set without loss of information, for example discarding two data points so that only two remain.

\begin{figure}
  \centering
  \subfigure[]
  {
    \includegraphics[scale=0.25]{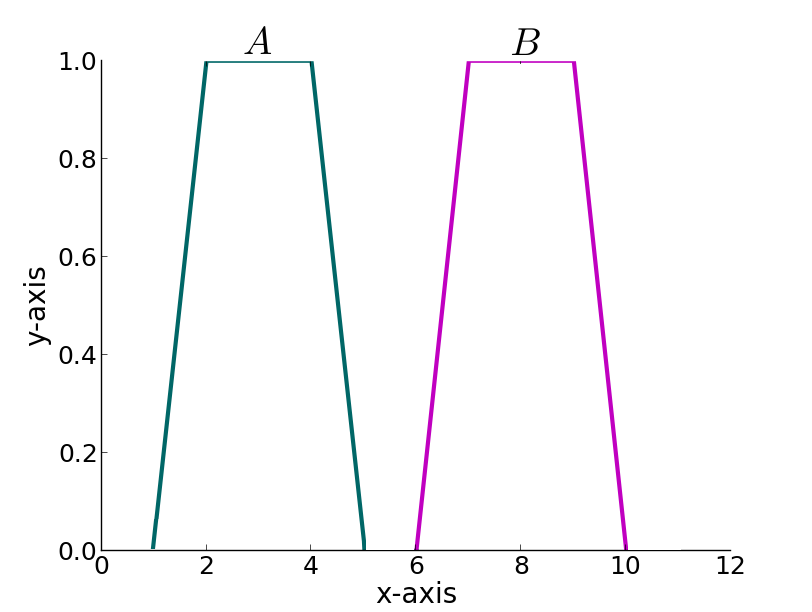}
  }
  \subfigure[]
  {
    \includegraphics[scale=0.25]{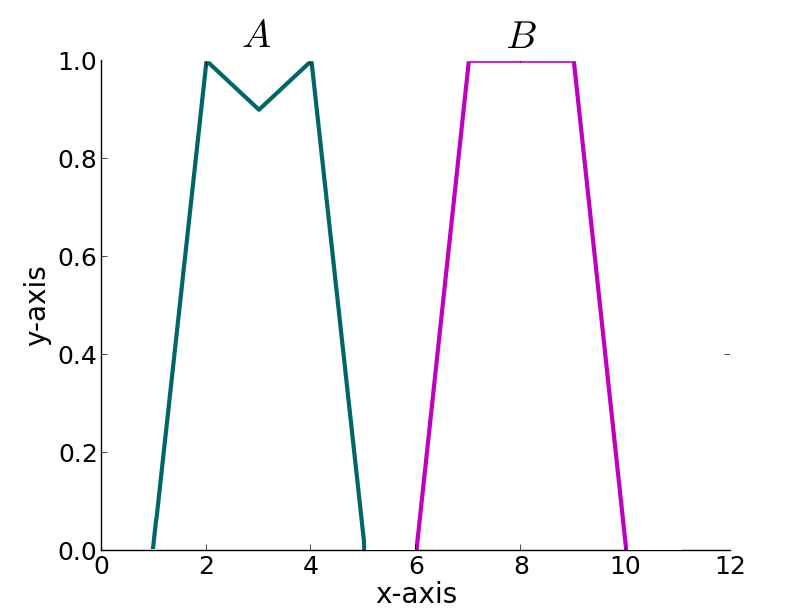}
  }
  \subfigure[]
  {
    \includegraphics[scale=0.25]{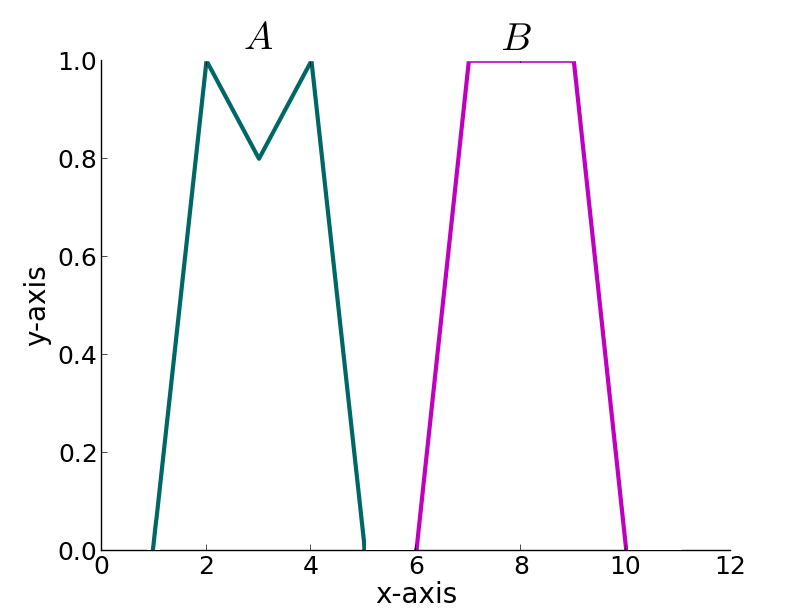}
  }
  \subfigure[]
  {
    \includegraphics[scale=0.25]{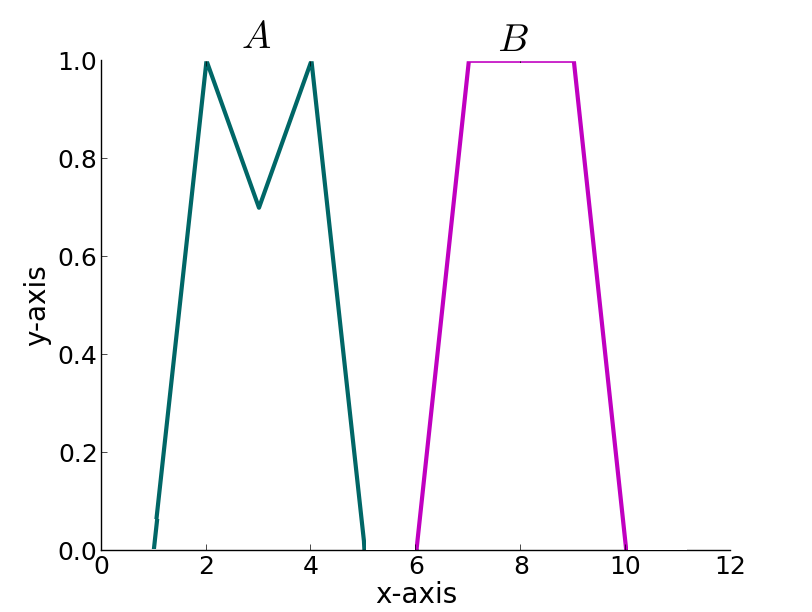}
  }
  \subfigure[]
  {
    \includegraphics[scale=0.25]{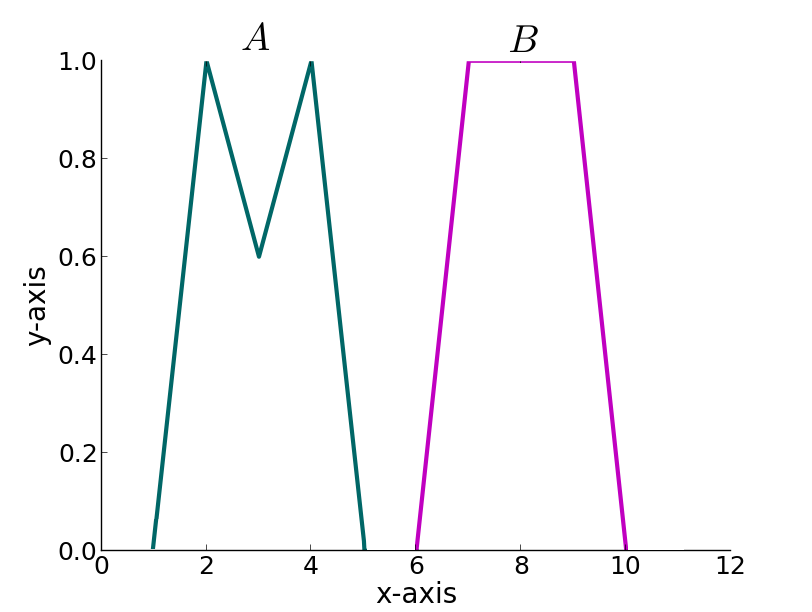}
    \label{fig:most_convex}
  }
  \caption{Comparing distance between a non-convex fuzzy set and a convex fuzzy set.}
  \label{fig:non_convex}
\end{figure}

Firstly, to solve this problem, it needs to be decided what the result of the distance measure should be. For example, in Fig. \ref{fig:non_convex}, should the distance increase as the fuzzy set $A$ becomes more concave, or should the distance decrease? It could be argued that since the similarity between the shapes of the two sets decreases, the distance should increase. The following proposes a method to solve this problem.

Fig. \ref{fig:most_convex} shows two fuzzy sets, $A$ and $B$. At $\alpha=0.8$, $A_\alpha = [1.8, 2.6, 3.5, 4.3]$, and $B_\alpha = [6.8, 9.2]$. One method to calculate the distance at this $\alpha$-cut is to split $A_\alpha$ into two intervals, $A_{\alpha 1}$ and $A_{\alpha 2}$. Next, we calculate the distance between $A_{\alpha 1}$ and $B_{\alpha}$ and the distance between $A_{\alpha 2}$ and $B_{\alpha}$. Then, to reduce the resulting two distances to one value, a fair approach is to take their average. Thus, the $\alpha$-cut at $\alpha=0.8$, is calculated as follows.

First, $A_\alpha$ is split into two intervals, $A_{\alpha 1} = [1.8, 2.6]$ and $A_{\alpha 2} = [3.5, 4.3]$. Next, using (\ref{eq:interval_haus_with_sign}), $h(A_{\alpha 1}, B_\alpha) = 6.6$ and $h(A_{\alpha 2}, B_\alpha) = 4.9$. Finally, the average of these is $5.75$ and is used as the result of $h(A_\alpha, B_\alpha)$ at $\alpha = 0.8$.

To test this method, an experiment was carried out on the sets in Fig. \ref{fig:non_convex} using this extension with RR's (\ref{eq:ralescu}) and CR's (\ref{eq:CR_haus}) measures. The non-convex set ($A$) was given as the first parameter of the distance measure, and the convex set ($B$) was given as the second parameter; thus, each measure should result in a positive distance. The results of the experiment are displayed in Table \ref{tab:non_convex_test}. These results show that as the fuzzy set $A$ becomes increasingly concave, the result of the distance measure increases.

Using this method, it is now possible, for example, to compare the ratings of the film \textit{All Dogs Go to Heaven 2} in Fig. \ref{fig:non_convex_film_set} with any of the films in \ref{fig:films_normalised}.

\setlength{\tabcolsep}{6pt}
\begin{table}[h!]
  \caption{Results of the extended Hausdorff measure applied to the fuzzy sets in Fig. \ref{fig:non_convex}; i.e. $d(A,B)$}
  \begin{center}
    \begin{tabular}{  c  c  c  c  c  c  }
      \toprule
        & a & b & c & d & e  \\ \midrule
        RR & 4.99 & 5.061 & 5.141 & 5.227 & 5.317 \\
        CR & 5.00 & 5.183 & 5.312 & 5.436 & 5.552
        \\ \bottomrule
    \end{tabular}
  \end{center}
  \label{tab:non_convex_test}
\end{table}

\section{Conclusions}
\label{sec:conclusions}
 In this paper we have introduced a distance measure for fuzzy sets which accounts for their direction of change, and have presented extensions of this distance measure for non-normal and non-convex fuzzy sets. We have demonstrated the applicability of the new distance measure using the MovieLens dataset and have asserted the advantage of using the new measure over distance measures which do not account for distance.
 
 In the future, we plan to further develop and test the new distance measures, and implement extensions for interval and general type-2 fuzzy sets \cite{mcculloch2013}. We will also apply the new distance measures within applications in CW, using fuzzy sets to construct word models of subjective information \cite{6251221} and apply distance-based reasoning.


\bibliographystyle{IEEEtran}
\bibliography{IEEEabrv.bib,papers.bib}

\begin{thebibliography}{10}
\providecommand{\url}[1]{#1}
\csname url@samestyle\endcsname
\providecommand{\newblock}{\relax}
\providecommand{\bibinfo}[2]{#2}
\providecommand{\BIBentrySTDinterwordspacing}{\spaceskip=0pt\relax}
\providecommand{\BIBentryALTinterwordstretchfactor}{4}
\providecommand{\BIBentryALTinterwordspacing}{\spaceskip=\fontdimen2\font plus
\BIBentryALTinterwordstretchfactor\fontdimen3\font minus
  \fontdimen4\font\relax}
\providecommand{\BIBforeignlanguage}[2]{{%
\expandafter\ifx\csname l@#1\endcsname\relax
\typeout{** WARNING: IEEEtran.bst: No hyphenation pattern has been}%
\typeout{** loaded for the language `#1'. Using the pattern for}%
\typeout{** the default language instead.}%
\else
\language=\csname l@#1\endcsname
\fi
#2}}
\providecommand{\BIBdecl}{\relax}
\BIBdecl

\bibitem{BonissoneLinguisticApproach80}
P.~P. Bonissone, ``{A fuzzy sets based linguistic approach: Theory and
  applications},'' in \emph{{Proceedings of the 12th conference on Winter
  simulation}}.\hskip 1em plus 0.5em minus 0.4em\relax IEEE Press, 1980, pp.
  99--111.

\bibitem{Wang20052063}
W.~Wang and X.~Xin, ``{Distance measure between intuitionistic fuzzy sets},''
  \emph{Pattern Recognition Letters}, vol.~26, no.~13, pp. 2063--2069, 2005.

\bibitem{analogicalReasoning}
I.~Turksen and Z.~Zhong, ``{An approximate analogical reasoning approach based
  on similarity measures},'' \emph{{IEEE} Trans. Syst., Man, Cybern.}, vol.~18,
  no.~6, pp. 1049--1056, 1988.

\bibitem{Xuecheng1992305}
L.~Xuecheng, ``{Entropy, distance measure and similarity measure of fuzzy sets
  and their relations},'' \emph{Fuzzy Sets and Systems}, vol.~52, no.~3, pp.
  305--318, 1992.

\bibitem{Zadeh1965}
L.~Zadeh, ``Fuzzy sets,'' \emph{Information and Control}, vol.~8, no.~3, pp.
  338 -- 353, 1965.

\bibitem{datamining}
K.~Hirota and W.~Pedrycz, ``Fuzzy computing for data mining,''
  \emph{Proceedings of the IEEE}, vol.~87, no.~9, pp. 1575 --1600, Sep. 1999.

\bibitem{timeSeries}
S.~Liao, T.~Tang, and W.-Y. Liu, ``Finding relevant sequences in time series
  containing crisp, interval, and fuzzy interval data,'' \emph{{IEEE} Trans.
  Syst., Man, Cybern. {B}}, vol.~34, no.~5, pp. 2071 --2079, Oct. 2004.

\bibitem{CWW}
L.~Zadeh, ``Fuzzy logic = computing with words,'' \emph{{IEEE} Trans. Fuzzy
  Syst.}, vol.~4, no.~2, pp. 103 --111, May 1996.

\bibitem{mendel2001uncertain}
J.~Mendel, \emph{{Uncertain rule-based fuzzy logic systems: introduction and
  new directions}}.\hskip 1em plus 0.5em minus 0.4em\relax Prentice Hall PTR,
  2001.

\bibitem{Zwick1987221}
R.~Zwick, E.~Carlstein, and D.~V. Budescu, ``{Measures of similarity among
  fuzzy concepts: A comparative analysis},'' \emph{International Journal of
  Approximate Reasoning}, vol.~1, no.~2, pp. 221--242, 1987.

\bibitem{Zadeh1975199}
L.~Zadeh, ``{The concept of a linguistic variable and its application to
  approximate reasoning---I},'' \emph{Information Sciences}, vol.~8, no.~3, pp.
  199--249, 1975.

\bibitem{Ralescu198485}
A.~L. Ralescu and D.~A. Ralescu, ``{Probability and fuzziness},''
  \emph{Information Sciences}, vol.~34, no.~2, pp. 85--92, 1984.

\bibitem{Chaudhur19961157}
B.~Chaudhur and A.~Rosenfeld, ``{On a metric distance between fuzzy sets},''
  \emph{Pattern Recognition Letters}, vol.~17, no.~11, pp. 1157--1160, 1996.

\bibitem{Fan1998793}
J.-l. Fan, ``{Note on Hausdorff-like metrics for fuzzy sets},'' \emph{Pattern
  Recognition Letters}, vol.~19, no.~9, pp. 793--796, 1998.

\bibitem{mcculloch2013}
J.~McCulloch, C.~Wagner, and U.~Aickelin, ``{Extending similarity mesaures of
  interval type-2 fuzzy sets to general type-2 fuzzy sets},'' in \emph{{Fuzzy
  Systems (FUZZ), 2013 IEEE International Conference on}}, 2013.

\bibitem{6251221}
S.~Miller, C.~Wagner, J.~Garibaldi, and S.~Appleby, ``{Constructing General
  Type-2 fuzzy sets from interval-valued data},'' in \emph{{Fuzzy Systems
  (FUZZ-IEEE), 2012 IEEE International Conference on}}, 2012, pp. 1--8.

\end{thebibliography}

\appendix[Numerical Example on Non-Normal Fuzzy Sets]
The proposed direction-based distance measure using (\ref{eq:proposed_non_normalised}) and (\ref{eq:interval_haus_with_sign}) is demonstrated using the non-normal fuzzy sets SMB and SW in Fig. \ref{fig:film_ratings_non_normalised}. The fuzzy sets for SMB and SW are distributed as follows:

\begin{equation}
  \nonumber
  \begin{array}{l @{\hspace{2bp}} l}
    SMB = & 0.385/1, 0.269/2, 0.231/3, 0.115/4, 0.0/5 \\
    SW\ =  & 0.015/1, 0.027/2, 0.098/3, 0.302/4, 0.557/5
  \end{array}
\end{equation}

Using linear interpolation, the $\alpha$-cuts of each film, given in the format $\alpha / [x_l, x_r]$, are:
\begin{equation}
  \nonumber
  \begin{array}{l @{\hspace{2bp}} l}
    SMB = & 0.1 / [1.0, 4.0], 0.2 / [1.0, 4.0], 0.3 / [1.0, 1.0] \\
    SW\  = & 0.1 / [2.44, 5.0], 0.2 / [3.08, 5.0], 0.3 / [3.36, 5.0], \\
          & 0.4 / [3.64, 5.0], 0.5 / [3.92, 5.0]
  \end{array}
\end{equation}

The $\alpha$-cut at 0.0 has been disregarded because it does not contribute to (\ref{eq:proposed_non_normalised}). The distance measure using (\ref{eq:proposed_non_normalised}) and (\ref{eq:interval_haus_with_sign}) is calculated as follows where $\lambda = 0.5$:\\
\begin{equation}
  \nonumber
  \begin{array}{l @{\hspace{2bp}} l}
    \text{At } \alpha & = 0.1 \text{, } h(SMB_\alpha, SW_\alpha) = 1.44 \\
    \text{At } \alpha & = 0.2 \text{, } h(SMB_\alpha, SW_\alpha) = 2.08 \\
    \text{At } \alpha & = 0.3 \text{, } h(SMB_\alpha, SW_\alpha) = 2.36 \\
    \text{At } \alpha & = 0.4 \text{, } h(SMB_\alpha, SW_\alpha) \\
	  & = h(A_{\alpha_k}, B_{\alpha_k}) \\
	  & = 2.36 \\
    \text{At } \alpha & = 0.5 \text{, } h(SMB_\alpha, SW_\alpha) \\
	  & = h(A_{\alpha_k}, B_{\alpha_k}) \\
	  & = 2.36
  \end{array}
\end{equation}
Finally, combining these results in (\ref{eq:proposed_non_normalised}) gives
\begin{equation}
 \nonumber
 \begin{array}{r @{\hspace{2bp}} l}
 d(A, B) & \\ 
    = & \frac{0.1 \times 1.44 + 0.2 \times 2.08 + 0.3 \times 2.36 + 0.4 \times 2.36 + 0.5 \times 2.36 }
             {0.1 + 0.2 + 0.3 + 0.4 + 0.5 } \\
    = & \frac{3.392}{1.5} \\
    = & 2.261
    
  \end{array}
\end{equation}

\end{document}